\pdfoutput=1
\documentclass[lettersize,journal]{IEEEtran}
\usepackage{amsmath,amsfonts}
\usepackage[ruled,vlined,linesnumbered]{algorithm2e}
\SetKwInput{KwData}{Data}
\SetKwInput{KwResult}{Result}

\usepackage{array}
\usepackage[caption=false,font=normalsize,labelfont=sf,textfont=sf]{subfig}
\usepackage{textcomp}
\usepackage{stfloats}
\usepackage{soul}
\usepackage{url}
\usepackage{verbatim}
\usepackage{graphicx}
\usepackage{amssymb}
\usepackage{cite}
\begin{document}

\title{Point Cloud-based Grasping for Soft Hand Exoskeleton}

\author{Chen Hu$^{1}$, Enrica Tricomi$^{2}$, Eojin Rho$^{3}$, Daekyum Kim$^{4}$, Lorenzo Masia$^{5}$, Shan Luo$^{1}$ and Letizia Gionfrida$^{1}$

\thanks{This work was supported by The Royal Academy of Engineering Research Fellowship (RF2324-23-229).
}
\thanks{$^{1}$C. Hu, S. Luo, and L. Gionfrida are with King's College London, London, WC2R 2LS, UK
        {\tt\small (email: \{tyrone.hu, shan.luo, letizia.gionfrida\}@kcl.ac.uk)}}%
\thanks{$^{2}$ E. Tricomi is with the Institut für Technische Informatik (ZITI), Heidelberg University, 69120 Heidelberg, Deutschland
{\tt\small (email: enrica.tricomi@ziti.uni-heidelberg.de)}}%
\thanks{$^{3}$E. Roh is with the School of Computing, KAIST, Daejeon 34141, South Korea 
{\tt\small (email: djwls9453@kaist.ac.kr).}}
\thanks{$^{4}$D. Kim is with the School of Mechanical Engineering and the School of Smart Mobility, Korea University, Seoul 02841, South Korea 
{\tt\small (email: daekyum@korea.ac.kr).}}
\thanks{$^{5}$ L. Masia is with the Munich Institute for Robotics and Machine Intelligence, Technical University of Munich, 80333 Munich, Deutschland
{\tt\small (email: lorenzo.masia@tum.de)}}%
}



\maketitle

\begin{abstract}
Grasping is a fundamental skill for interacting with and manipulating objects in the environment. However, this ability can be challenging for individuals with hand impairments. Soft hand exoskeletons designed to assist grasping can enhance or restore essential hand functions, yet controlling these soft exoskeletons to support effectively users remains difficult due to the complexity of understanding the environment. This study presents a vision-based predictive control framework that leverages contextual awareness from depth perception to predict the grasping target and determine the next control state for activation. Unlike data-driven approaches that require extensive labelled datasets and struggle with generalizability, our method is grounded in geometric modelling, enabling robust adaptation across diverse grasping scenarios. The Grasping Ability Score (GAS) was used to evaluate performance, with our system achieving a state-of-the-art GAS of 91 ± 2\% across 15 objects and healthy participants, demonstrating its effectiveness across different object types. The proposed approach maintained reconstruction success for unseen objects, underscoring its enhanced generalizability compared to learning-based models.
\end{abstract}


\section{Introduction}
\IEEEPARstart{I}{ndividuals} with grasping impairments can experience restriction of hand function \cite{raghavan2007nature}. For instance, individuals with spinal cord injury or post-stroke lose the ability to extend their fingers, but retain unaffected motor and sensory abilities outside the hand \cite{Silver2021}. Such impairments make it nearly impossible to perform essential grasping tasks, posing significant challenges to independence and quality of life \cite{raghavan2007nature}.

Over the past decade, soft hand exoskeletons have been developed to assist users in performing daily grasping tasks \cite{gionfrida2024wearable} by providing additional force to users' fingers to support grasping activities \cite{du2021review}. Unlike rigid hand exoskeletons, which use 3D-printed from stiff materials to immobilize fingers and execute predefined movements, soft hand exoskeletons \cite{kim2019eyes} are made of flexible materials such as fabric and silicone, offering enhanced comfort during use. 

Current soft hand exoskeletons enable assistance through pneumatic-based \cite{ge2020design} or tendon-driven \cite{rho2021learning} mechanisms to facilitate finger flexion and extension, thereby aiding users in grasping and releasing objects. Tendon-driven soft hand exoskeletons have garnered attention due to their ability to mimic the natural movement patterns of tendons during object grasping and releasing\cite{coyle2018bio}.
A challenge for such exoskeletons involves determining how to effectively regulate control based on user intention \cite{gionfrida2024wearable,tricomi2023environment}. Many such devices rely on surface electromyography (sEMG) \cite{de2022emg}, detecting muscle electrical activity to compensate for residual muscle activity \cite{siviy2023opportunities}. 

The integration of visual perception into the control strategies of wearable robots has gained increasing attention in the robotics community \cite{gionfrida2024wearable}, offering the potential to enhance adaptability and decision-making in dynamic environments. 

Current state-of-the-art methods predominantly rely on data-driven approaches that incorporate deep learning-based visual perception \cite{tricomi2023environment, kim2019eyes}. While these techniques demonstrate strong performance in controlled settings, their application in wearable robots is hindered by three key factors: (1) the requirement for extensively trained vision algorithms, (2) the lack of task-specific datasets tailored to wearable robotic applications, and (3) the computational cost associated with processing high-dimensional visual data in real-time. These constraints pose significant challenges in seamlessly integrating visual feedback into wearable robotic systems.

An alternative to data-driven methods is to leverage geometric modelling for scene understanding \cite{diaz2016strong}, which reduces the reliance on large-scale annotated datasets. Depth-based object modelling, for instance, allows the inference of scene geometry \cite{hu2024pointgrasp}. Unlike learned feature representations, which are often sensitive to changes in camera viewpoints, geometric modelling exploits the underlying spatial structure of the environment, making it inherently more adaptable to unseen scenarios \cite{perez2015detection}. By shifting the computational focus from high-cost inference to geometric reasoning, these methods can enable real-time visual processing, enhancing responsiveness in wearable robotics \cite{driess2021learning}. Integrating geometric modelling into vision-based control strategies can improve robustness, adaptability, and efficiency, offering a scalable alternative to deep learning-based solutions for real-world assistive applications. \\

\begin{figure*}[t]
\centering
\includegraphics[width=1\textwidth]{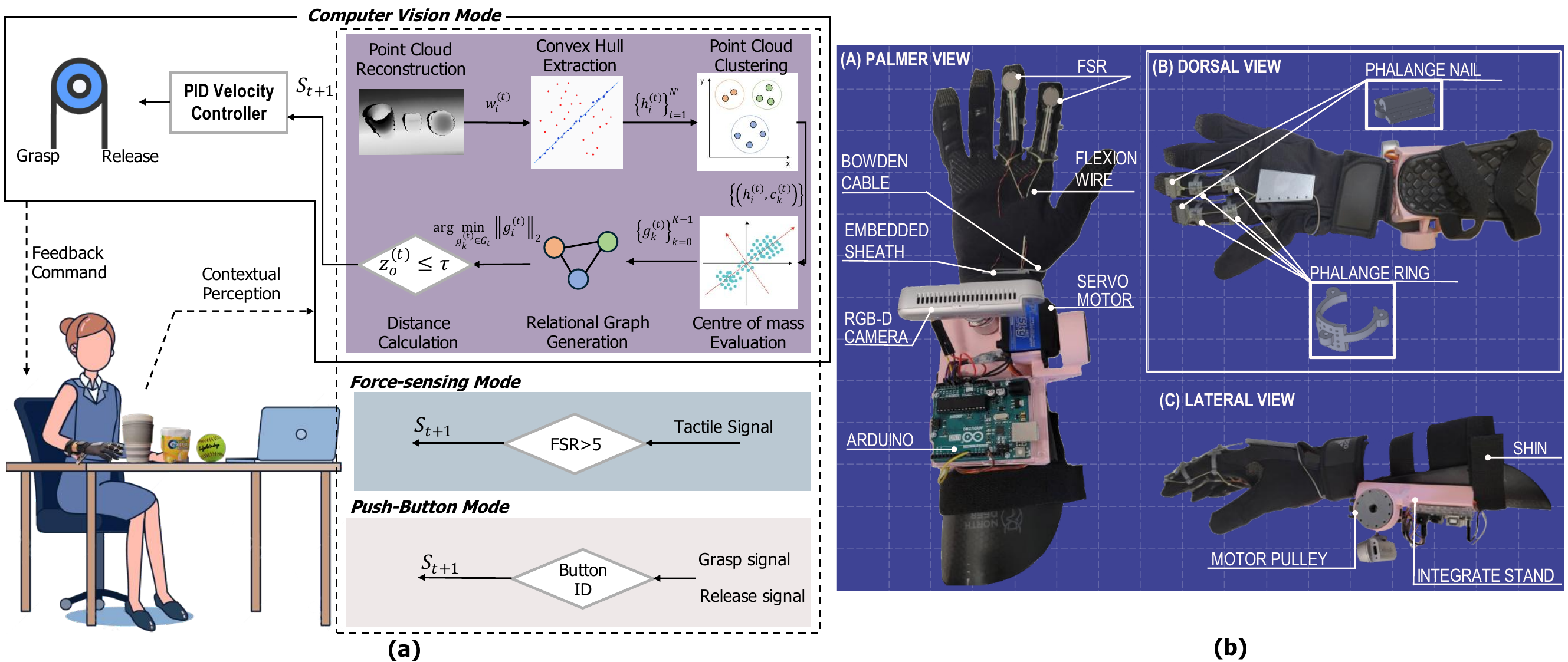}
\caption{System design: \textbf{(a)} A vision-based controller was developed for the study. The controller reconstructs the 3D point cloud from depth frames using the camera's intrinsic parameters. Neighbourhood density is calculated as a confidence measure to sort the point cloud, and the largest planar model (table) and the convex hull (objects) in the contextual perception are identified using PROSAC \cite{chum2005matching}. The category of each point in the point cloud is determined using DBSCAN \cite{Ester1996DBSCAN}. Principal Component Analysis (PCA) \cite{Pearson1901PCA} is applied to compute the centroid of each object category. These centroids are used to generate a relationship graph among objects, and the centroid closest to the camera's optical axis is identified as the target object. When the distance between this object and the camera plane is less than an adaptable threshold $\tau$, velocity PID control is triggered to assist the user in completing the grasping task. \textit{Force-sensing mode:} a force-sensitive resistor (FSR) sensor mounted at the fingertip of the hand exoskeleton measures pressure values; a grip command is triggered when the pressure from either sensor exceeds a defined threshold. \textit{Push-button mode}: commands are transmitted via the corresponding button presses. \textbf{(b)} The soft hand exoskeleton, based on an existing design \cite{rho2021learning}, consists of three main components: an embedded actuator, a customized soft exoskeleton that transfers force to the finger joints, and a sensing module. The actuation system is mounted on a shin guard worn on the forearm for optimized weight distribution. Additionally, 3D-printed nails and rings are designed and installed near the metacarpophalangeal, proximal interphalangeal, and distal interphalangeal joints to replicate flexion tendons passing through the joints.
}
\label{fig1_hand exoskeleton_overview}
\end{figure*}
Geometric point cloud has also been used to detect grasp points for robotic grippers \cite{duan2021robotics}, but this approach has not yet been explored for wearable robots. Although individuals without impairments can determine grasp points intuitively, those with impairments, may struggle to identify when and where to grasp due to cognitive challenges \cite{el2023cognitive}. Point detection can assist those who approach a target object but struggle to determine how to initiate the grasp, enhancing stability, functionality, and the overall effectiveness of the assistance. 

Building on these insights, this work introduces a geometric vision-based control system for a soft hand exoskeleton. The proposed framework integrates a wrist-mounted depth camera with a soft wearable robotic hand exoskeleton (Fig. 1). By leveraging geometric point cloud, the system reconstructs the 3D scene to identify grasp points and actuates the tendon-driven exoskeleton to execute hand closure. The approach is validated through experiments, including performance comparisons with existing force-sensing and push-button control modalities and tests involving healthy participants performing diverse grasping tasks with different objects. The key contributions of this work include:
\vspace{-0.1cm}
\begin{enumerate}
\item A geometric point cloud framework to improve generalizability and computational speed compared to traditional data-driven vision controllers.

\item A vision-based controller to detect grasping points that improve the grasping ability score in comparison to other control approaches.
\end{enumerate}

\section{Methodology}
The proposed vision-based control framework (Fig. \ref{fig1_hand exoskeleton_overview} \textbf{(a)}) was evaluated using a tendon-driven soft hand exoskeleton (Fig. \ref{fig1_hand exoskeleton_overview} \textbf{(b)}), based on an existing design \cite{rho2021learning}, and tested across different objects and grasp types against two existing control modes: force-sensing \cite{rho2021learning}, and push-button \cite{alicea2021soft}. 
\subsection{Dataset}

The dataset comprises 15 objects, including 7 (banana, strawberry, softball, apple, pear, orange, and plum) selected from the Yale-CMU-Berkeley (YCB) dataset \cite{calli2015benchmarking}, designated as \textit{'seen'} objects, and 8 additional everyday objects (chewing gum box, small storage box, purse, small chips can, cup, coffee can, peach can, and chilli can), classified as \textit{'unseen'}, chosen to evaluate the generalizability of the proposed algorithm. 
Items were chosen to be classified into three distinct grasp types: pinch, spherical grip, and cylindrical grip, as in the study proposed by Calli et al. \cite{calli2015benchmarking}, ensuring heterogeneity across the objects in aspects such as shape, mass, dimensions, and material composition.


\begin{figure}[t]
\centering
\includegraphics[width=0.45\textwidth]{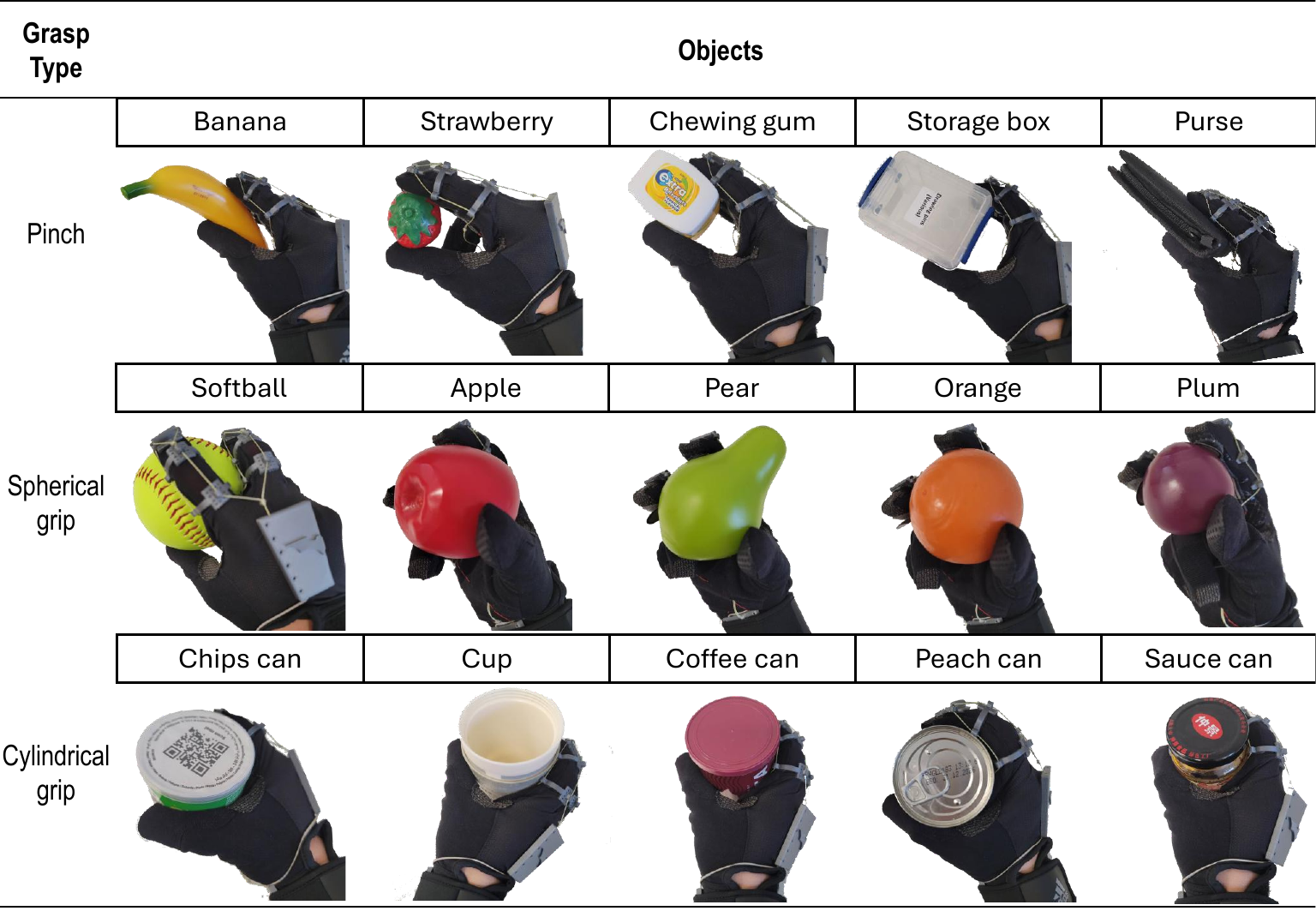}
\vspace{-0.3cm}
\caption{The three grip types, including pinch, spherical grip, and cylindrical grip, are used to grasp 15 objects. 
}
\label{plot_grasp_types}
\vspace{-0.55cm}
\end{figure}


\subsection{Hardware Design}

The implemented soft hand exoskeleton used in the study was based on an existing design \cite{rho2021learning}, and consisted of three main components (Fig. \ref{fig1_hand exoskeleton_overview} (b)): 1) a Tendon-Sheath Mechanism (TSM) actuator; 2) a custom glove that transferred force to the finger joints; and 3) a sensing module.

The actuation system, powered by a LiPo battery (Crazepony, 1400mAh, 11.1V, 64.1g, Shenzhen, China), weighed 0.5 kg and was mounted on a forearm-worn shin guard (Super Comfortable Shin Pad, Northdeer, China). A flat servo motor (Digital Servo, 4.8V, 24 kgcm, CHICIRIS, China) drove a 30mm diameter pulley, around which the actuation cable was wound. The system employed a microcontroller (Arduino Uno, Ivrea, Italy) to manage force-sensing analog-to-digital conversion and motor pulse-width modulation (PWM).

The exoskeleton was composed of flexible polyamide, with 3D-printed rings near the metacarpophalangeal (MCP), proximal interphalangeal (PIP), and distal interphalangeal (DIP) joints, replicating human hand tendons for effective grasping and releasing motions. A TSM connected the actuation system to the glove and was designed to remain fixed, reducing the impact of dynamic sheath bending angles. To maintain a tight, straight connection between the actuation system and the palm-supporting pieces, the sheaths were slightly tensioned, which minimized sheath bending during operation. This approach reduced unwanted contact between the tendon wires and the sheath, effectively stabilizing fingertip force against dynamic changes in the sheath’s bending angle.

The sensing module, following the existing design \cite{rho2021learning}, incorporated a force-sensing resistor (FSR) sensor (Oumefar, China), integrated into an RC circuit, with a resistance value of \( R_{M} \) set at 10 k\(\Omega\) and an input voltage \( V_{+} \) of 5V. The circuit was mounted on the tip of the index finger and an actuation command was activated when the pressure exceeded a defined threshold. 
To introduce and evaluate the vision-based framework, a wrist-mounted RealSense D415 RGB-D camera (Intel RealSense, California, USA) was incorporated into the soft exoskeleton original design, to capture environmental data. The camera, is attached to the shin pad via a miniature gimbal. sent depth images are sent to a server (Dell, Precision 7680, US) for inference with a 10 frames per second (fps)  frame rate and 640 $\times$ 480 resolution.
To assess the effectiveness of the vision-based approach, apart from the existing  force-sensing mode, the sensing module incorporated an additional control modality: a push-button \cite{alicea2021soft}, where actuation commands were transmitted via the corresponding
button presses.

\begin{algorithm}[t]
\caption{Proposed vision-based control approach}
\label{alg:vision_control}
\KwData{Depth matrix $M_t(u, v)$, camera intrinsic matrix $\Phi$, parameters $\epsilon$, $\tau$, $m_0$, $\mu$}
\KwResult{Estimated PID control state $S_{t+1}$}

\While{not\ thread\_stop}{
  \If{$M_t(u, v)$ exists}{
    $P \gets \{p_i^{(t)} = (x_i^{(t)}, y_i^{(t)}, z_i^{(t)})\}_{i=1}^N$ from $M_t(u, v)$ and $\Phi$

    $N_\epsilon(p_i^{(t)}) \gets \{p_j^{(t)} \mid \|p_i^{(t)} - p_j^{(t)}\| \leq \epsilon\}$

    $\rho_i^{(t)} \gets |N_\epsilon(p_i^{(t)})|$

    $w_i^{(t)} \gets \frac{\rho_i^{(t)} - \rho_{\min}^{(t)}}{\rho_{\max}^{(t)} - \rho_{\min}^{(t)}}$

    $P \gets \text{Sort}(P, w_i^{(t)})$

    $m(0) \gets m_0$

    \Repeat{plane model $\pi^*$ is found with sufficient support}{
        Select top $m(n)$ points with highest $w_i^{(t)}$

        Fit plane $\pi$: $Ax + By + Cz + D = 0$

        Update $m(n+1) \gets \min(N, m_0 + n)$
    }

    $H_t \gets P \setminus \pi^*$

    $clusters \gets \text{DBSCAN}(H_t, \mu)$

    $G^* \gets \arg\min_{G_k} \|c_k^{(t)}\|_2$

    $distance\_to\_camera \gets \|c_{G^*}^{(t)}\|_2$

    \If{motor\_state == \text{True}}{
      \eIf{$distance\_to\_camera < \tau$}{
          $S_{t+1} \gets v_{t+1}$
      }{
          $S_{t+1} \gets 90$
      }
    }
  }
}

\end{algorithm}

\subsection{Control Design}

\subsubsection{Vision-Based control mode}

The vision-based process began with the reconstruction of a 3D point cloud from depth frames. The neighbourhood density was then calculated as a confidence measure to sort the point cloud. Next, the largest planar model (representing the table where the objects were placed) and the convex hull (representing objects) in the contextual perception were identified using the PROSAC \cite{chum2005matching}. PROSAC used a progressive priority sampling strategy which, in contrast to random sampling, began with high-confidence points and gradually expanded to lower-confidence ones. This approach enabled the correct model to be identified with significantly fewer iterations \cite{chum2005matching}. 
To categorize the points in the cloud, the DBSCAN \cite{Ester1996DBSCAN} was applied. Unlike traditional clustering algorithms, DBSCAN did not require a predefined number of clusters, making it well-suited for analyzing uncertain and dynamic environments. Its density-based clustering process enabled the identification of irregular and arbitrarily shaped structures in point cloud data, providing a robust framework for unstructured point cloud analysis \cite{ahmed2020density}. 
Principal Component Analysis (PCA) \cite{Pearson1901PCA} was then used to compute the centroid of each object. These centroids were utilized to generate a relationship graph among objects, and the centroid closest to the camera’s optical axis was selected as the target object. When the distance between the object and the camera plane was less than a threshold, a PID control was triggered to assist the user in completing the grasping task. Due to the visible range of the RealSense D415, the threshold was from 200 mm to 10000 mm. Its adaptability allowed for the automated selection of an appropriate threshold based on the object's size and the hand's dimensions. For example, using a caliper, we measured the grasping distance for a banana across 30 trials, yielding an average distance of approximately 400 millimeters (mm).

The essence of the proposed control method, $\mathcal{F}$, was to predict the grasping target based on the current depth frame to determine the velocity PID control state at the next moment. 
We established the following mathematical model:
\begin{equation}
S_{t+1}=\mathcal{F}(M_t(u,v); \Phi, \epsilon, \tau,  m_0, \mu)
\label{equ:1}
\end{equation}

\noindent where $M_t(u,v)$ is the depth matrix obtained by the camera at time $t$, $u$ and $v$ are the depth matrix coordinates, $\Phi$ is the depth camera intrinsic matrix, $\epsilon$ is the neighbourhood radius, $\tau$ is the PROSAC \cite{chum2005matching} threshold for determining the maximum plane model's error value, $m_0$ is the initial sampling range size, $\mu$ is the minimum number of points required by DBSCAN \cite{Ester1996DBSCAN} to evaluate core points, and $S_{t+1}$ is the velocity PID controller state at time $t+1$. The intrinsic matrix of the RealSense camera is defined as:
\begin{equation}
\boldsymbol{\Phi} = \begin{bmatrix} f_x & 0 & c_x \\ 0 & f_y & c_y \\ 0 & 0 & 1 \end{bmatrix}
\label{equ:2}
\end{equation}

Using Open3D \cite{Zhou2018Open3D}, we reconstructed the depth matrix $M_t(u,v)$ into a point cloud $P=\{p_{i}^{(t)}\}_{i=1}^{N}$ (Fig. 1(a)), where $N$ is the total number of points in the point cloud. The point cloud reconstruction outputs are visualized for 15 objects using the proposed vision-based approach (Fig. \ref{plot_point_cloud_detection}) to illustrate the output of the method.

Any point in this dataset was defined as $p_{i}^{(t)}=\{x_i^{(t)},y_i^{(t)},z_i^{(t)}\}$. Each point's coordinates in the camera coordinate system were calculated as follows:
\begin{equation}
Z=M_t(u,v), X=\frac{(u - c_x) \cdot Z}{f_x}, Y=\frac{(v - c_y) \cdot Z}{f_y}
\label{equ:3}
\end{equation}

For any point $p_{i}^{(t)}$ in the scene point cloud, its neighborhood set $N_{\epsilon}(p_{i}^{(t)})$ was defined as:
\begin{equation}
N_{\epsilon}(p_{i}^{(t)}) = \{ p_{j}^{(t)} \in P_t \mid \| p_{i}^{(t)} - p_{j}^{(t)} \| \leq \epsilon \}
\label{equ:4}
\end{equation}

\noindent where $\| p_{i}^{(t)} - p_{j}^{(t)} \|$ is the Euclidean distance between $p_{i}^{(t)}$ and $p_{j}^{(t)}$. The neighborhood density of $p_{i}^{(t)}$ was defined as $\rho_i^{(t)} = |N_{\epsilon}(p_{i}^{(t)})|$. Based on the density $\rho_i^{(t)}$, the confidence of $p_{i}^{(t)}$ was defined as:
\begin{equation}
w_i^{(t)} = \frac{\rho_i^{(t)} - \rho_{\text{min}}^{(t)}}{\rho_{\text{max}}^{(t)} - \rho_{\text{min}}^{(t)}}
\label{equ:5}
\end{equation}

Here, $w_i^{(t)} \in [0, 1]$, and $\rho_{\text{min}}^{(t)}$ and $\rho_{\text{max}}^{(t)}$ are the minimum and maximum neighbourhood densities among all points. Sorting points by confidence $w_i^{(t)}$ in descending order results in an ordered point set. The top $m_{0}$ points with the highest confidence are selected as the initial sampling range. These points were considered most likely to belong to the target plane. In the $n$-th iteration, the sampling range was progressively expanded, and the expansion was expressed as:
\begin{equation}
m(n) = \min(N, m_0 + n)
\label{equ:6}
\end{equation}

\noindent where $m(n)$ represents the sampling range size in the $n$-th iteration. The sampling range in the $n$-th iteration is denoted as $P'_{t}$. We randomly select three points from $P'_{t}$ to fit a plane model $\{ P'_a, P'_b, P'_c \} \subseteq P'_t$. Early iterations used smaller sampling ranges $m(n)$, containing only the highest-confidence points, which were more likely part of the plane. As iterations proceed, the range gradually expands to include more points, ultimately encompassing all points. This ensures that points with initially low confidence but actually part of the plane can be included. Using $\{ P'_a, P'_b, P'_c \}$, the plane normal vector is calculated as:
\begin{equation}
\vec{l} = (P'_b - P'_a) \times (P'_c - P'_a) = (A, B, C)
\label{equ:7}
\end{equation}

This gives the plane equation $Ax+By+Cz+D=0$.
The distance from point $p_{i}^{(t)}$ to the plane model $\pi$ is expressed as:
\begin{equation}
Dist{(p_{i}^{(t)},\pi)} = \frac{|A x_i^{(t)} + B y_i^{(t)} + C z_i^{(t)} + D|}{\sqrt{A^2 + B^2 + C^2}}
\label{equ:9}
\end{equation}

Based on the above formula, the distance of each point to the plane $\pi$ can be calculated. The support of the plane is evaluated as $I=\sum_{i=1}^{N}1(Dist{(p_{i}^{(t)},\pi)})$.

After all points were processed, the plane model with the maximum support was selected as the maximum plane model $\pi^*$. The complement of the maximum plane model was obtained to get the convex hull point cloud, denoted as: $H_t=\{h_i^{(t)}\}_{i=1}^{N'}$,
\noindent where $H_t \subseteq P_t$ and $N' < N$. According to DBSCAN, all points were classified into core points, boundary points, and noise points. A core point must satisfy $N_{\epsilon}(h_{i}^{(t)})\geq\mu$; a boundary point satisfies $N_{\epsilon}(h_{i}^{(t)}) < \mu$ but is within the neighborhood of a core point; a noise point satisfies $N_{\epsilon}(h_{i}^{(t)}) < \mu$ and is not in any core point’s neighborhood. DBSCAN first traverses $H_t$ to identify all core points. Then, starting from any unvisited core point $h$, a new cluster was created. All points within the $\epsilon$ neighbourhood of $h$ were added to the cluster. If these points included core points, their neighbourhoods were added to the cluster. Boundary points were automatically added to the nearest core point’s cluster. This process repeats until the current cluster no longer expands. Unclustered points were marked as noise. Once all points in $H_t$ were processed, a set of clusters with $K$ categories was obtained: $C_t=\{c_k^{(t)}\}_{k=0}^{K-1}$, where $i=-1$ represents noise points. After removing noise points, the point-class relationship was as: $\{(h_{i}^{(t)},c_{k}^{(t)})|h_i^{(t)} \in H_t,c_k \in \{0,...,K-1\}\}$.
Using PCA \cite{Pearson1901PCA}, we computed the centroids of each cluster to represent the scene graph of objects on the table at time $t$: $G_t=\{g_k^{(t)}\}_{k=0}^{K-1}$.
The node in $G_t$ closest to the camera optical axis $Z$ was selected as the target object:
\begin{equation}
\vspace{-0.1cm}
(x_o^{(t)},y_o^{(t)},z_o^{(t)})=\arg\min_{\mathbf{g}_i^{(t)} \in \mathcal{G_t}} \|\mathbf{g}_k^{(t)}\|_2
\label{equ:14}
\vspace{-0.1cm}
\end{equation}

If the distance between the target object and the camera plane at time $t$ was greater than or equal to a threshold $\tau \in (200, 10000)$, measured in mm, the velocity PID at time $t+1$ was set to 90, keeping the motor stationary. If the distance is less than or equal to $\tau$, the velocity PID at time $t+1$ was set to $v_{t+1}$, making the motor rotate at a constant speed of $v_{t+1}$, driving the exoskeleton to continuously close.
\begin{equation}
S_{t+1} = \begin{cases} 90, & \text{if } z_o^{(t)} \geq \tau \\ v_{t+1}, & \text{if } z_o^{(t)} < \tau \end{cases}
\label{equ:15}
\vspace{-0.1cm}
\end{equation}

\subsubsection{Force-sensing control mode}

The FSR sensor fixed on the tip of the index finger of the glove, following the original design proposed by \cite{rho2021learning}, was used to initiate grasping. This configuration initiated the grasp command when the fingertip sensors came into contact with an object. The motor received a grip command to rotate forward and output a constant torque of 24 kgcm ($\approx$ 2.35 Nm) when the power system provided 4.8V voltage, assisting the user in grasping and maintaining actions. Based on the benchmarking comparison, this precise torque aided users in grasping objects. According to \cite{polygerinos2015soft}, the required force to grasp objects during daily activities (ADL) does not exceed 15 N, and the pinch forces required to execute most of the daily life tasks are lower than 10.5 N \cite{smaby2004identification}. The FSR was also connected in series with a 3 \text{k}$\Omega$ resistor $R_{M}$ and referenced to a 5V source $V_{+}$, with the voltage drop measured by the 32-bit Register ADC of an ESP32 controller. The output voltage $V_{OUT}$ increased with increasing force. The relationship between force and voltage under different resistances was derived from the Seeed Studio 101020553 Datasheet \cite{datasheet}
. As the grasping motion commenced upon activation of the pressure sensor in contact with the object, a small pressure dead zone was established at a threshold of $\Delta 5$ ADC readings to prevent unintended actions.


\subsubsection{Push-button control mode}

Another common control mode was implemented, where the soft hand exoskeleton control was triggered by pressing a button, as in \cite{alicea2021soft}. When the user needed to grasp an object, they approached the object with their right hand and pressed the grasp button with their left hand. This triggered the motor to rotate forward, delivering a constant torque of 24 kgcm. A PID controller adjusted the phase currents to maintain these parameters, enabling the object to be grasped and maintained stably. Similarly, the user could press the release button to trigger the motor to rotate backwards, resulting in a releasing action.

\subsection{Study Protocol and Evaluation}

To assess the grasping performances of the proposed vision-based control architecture against the force-sensing and push-button modes, 10 right-handed healthy participants were involved in the study: seven males and three females aged 25.0±6 years, weighing 75±14 kg, measuring 1.80±0.22 m in height. All participants demonstrated standard hand motor functions. Participants were instructed to execute grasping tasks, ensuring that these actions were performed without any discomfort or pain. The study received ethical approval (\textit{MRPP-23/24-40750}) from the College Research Ethics Committee at King’s College London. Information detailing the study’s aims and procedures was provided to all participants, who subsequently gave their written informed consent before involvement.

Once they agreed to participate, participants sat adjacent to a table, as sketched in Fig. \ref{fig1_hand exoskeleton_overview} (a). Participants had to keep their hands relaxed and avoid applying force on the object during the grasp. A researcher handed the objects to the participants, who were instructed to maintain their grasp for three seconds. Subsequently, participants were asked to rotate their hand 180° to a palm-down position, holding the grasp for another three seconds before the system initiated the release. Moreover, the researcher recorded how many times the camera came into contact with the table to estimate whether placing this sensor underneath caused discomfort for the users. 
\subsubsection{Grasping Ability Score}
Following the protocol proposed by \cite{maldonado2023fabric}, each object was grasped three times in each control mode to test the proposed vision-based approach's capability to execute the grasping task (Grasping) and its ability to hold the object (Maintaining). The Grasping and Maintaining scores were assessed from the moment the object was attempted to be grasped until it was released. The Grasping score was assigned values of 0, 0.5, and 1, with 0 for failure, 0.5 for an incorrect hand grasp, and 1 for the correct grasp type. Similarly, the Maintaining score was rated from 0 to 1, with 0 assigned if the object was dropped, 0.5 if there was movement, and 1 if the object remained stable. The scores from the objects within each grasp type were combined to derive a final score for each grasp. A cumulative score, the Grasping Ability Score (GAS) \cite{llop2019anthropomorphic}, was then calculated to represent the vision-based approach's overall performance in executing the grasps. 


\begin{figure*}[t]
\centering
\includegraphics[width=0.85\textwidth]{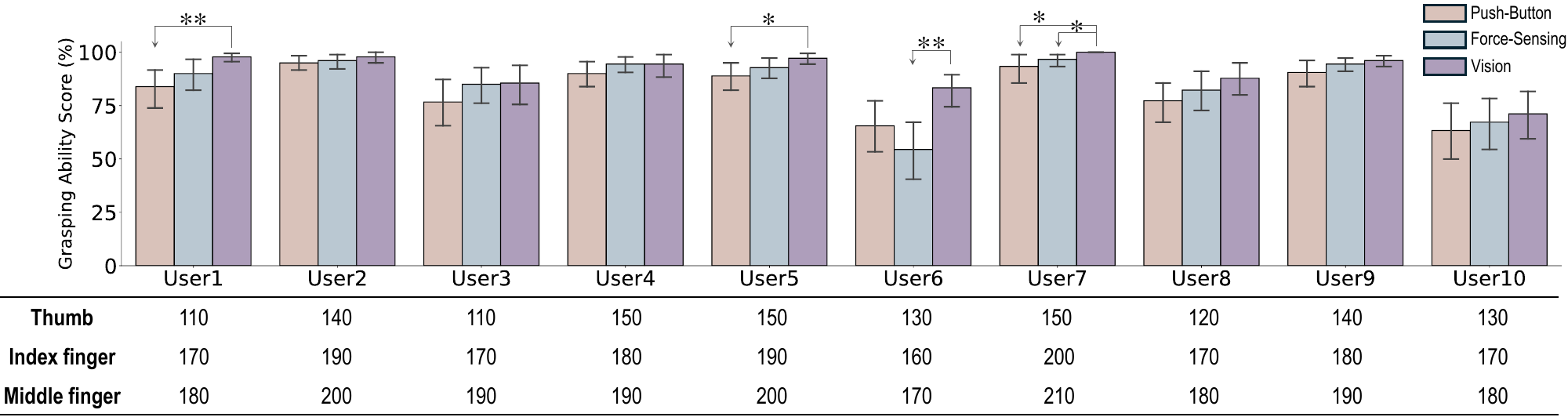}
\caption{Grasping Ability Scores (GAS) in percentage (\%) for 10 users when grasping 15 objects across three control modes: push-button mode (orange), force-sensing mode (blue), and vision-based mode (purple) are displayed in the bar chart. The table below shows the distances (mm) between the tips of the thumb, index finger, middle finger, and wrist. Longer fingers correlate with higher GAS. Significance levels are indicated by asterisks (*), where * denotes p $\leq$ 0.05, ** denotes p $\leq$ 0.01, and *** denotes p $\leq$ 0.001.}
\label{plot_gas_user}
\vspace{-0.3cm}
\end{figure*}

\begin{figure*}[t]
\centering
\includegraphics[width=1\textwidth]{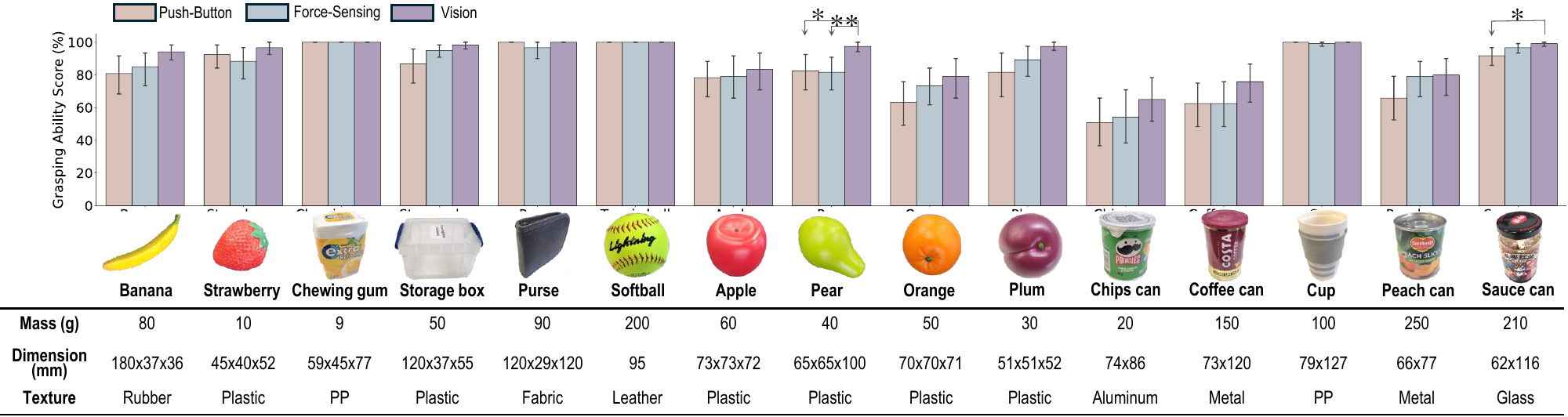}
\vspace{-0.5cm}
\caption{Grasping Ability Scores (GAS) for 15 objects across three grasping modes: the push-button mode (orange), force-sensing mode (blue), and vision-based mode (purple),  displayed in a bar chart. The table shows the parameters of 15 objects, including mass, dimensions, and material to illustrate their influence on the GAS. Significance levels are indicated by asterisks (*), where * denotes p $\leq$ 0.05, ** denotes p $\leq$ 0.01, and *** denotes p $\leq$ 0.001.}
\label{plot_gas_object}
\vspace{-0.5cm}
\end{figure*}


\subsubsection{Vision-based mode vs. data-driven approaches}

Performance of the proposed visual method over data-driven approaches were evaluated from two key perspectives: response time, and generalizability across \textit{unseen} objects. For data-driven approaches, pre-trained YOLO11n-seg and YOLO11x-seg models \cite{yolo11_ultralytics} were used. We selected YOLO11-seg for instance segmentation of objects within the scene, followed by 3D reconstruction of the target object using Open3D \cite{Zhou2018Open3D}. 

As the user approached the object, we extracted keyframes every 20 frames from the first 100 frames, resulting in a total of 5 keyframes. For these keyframes, we calculated the reconstruction success rates ($RSR$)  as: 
\begin{equation}
RSR = \frac{Frame\_Number_{success\_reconstruction}}{Total\_Frame\_Number}
\label{equ:16}
\end{equation}

\subsubsection{Kinematics analysis of fingers}
The biomechanical impact of the proposed controller on the range of motion (ROM) was evaluated using video recordings captured during the experiment. The ROM was measured at key joints— MCP, PIP, and DIP—during grasping tasks performed with and without the hand exoskeleton across three grasp types: cylindrical (C), spherical (S), and pinch (P). To extract the ROM, YOLO11 \cite{yolo11_ultralytics} was fine-tuned on the 11k Hands Dataset \cite{afifi201911kHands} over 100 epochs. The output was then analyzed to compare the ROM with and without external actuation, providing an assessment of the proposed controller's performance.

\section{Results and Discussion}
\subsection{Grasping Ability Score (GAS)}
Fig. \ref{plot_gas_user} and Fig. \ref{plot_gas_object} present the GAS \cite{llop2019anthropomorphic} in percentage (\%) across 10 participants and 15 objects, respectively. The results are reported for the three modes: vision-based, force-sensing, and push-button. P-values are categorized into four levels based on their significance. On average, the vision-based mode demonstrates slightly superior successful grasping and maintaining performances (Fig. \ref{plot_gas_user}), with an average GAS across participants of 91±2\%, 6\%, and 9\% higher than the force-sensing and push-button modes, respectively. The slight improvement can be attributed to the system’s activation mechanism. Particularly, when the depth camera detects the object’s centroid at a distance of less than an adaptable threshold $\tau$, a 3-second delay is triggered before the soft exoskeleton closes. This delay allows users additional time to approach the object, possibly leveraging their cognitive abilities to ensure a successful grasp, as in \cite{llop2019anthropomorphic}. 


The GAS analysis for 15 different objects in three distinct grasping modes shows that the grasp success rate can be influenced by the surface material, size and mass of the objects (Fig. \ref{plot_gas_object}). The object's texture affects the ability to achieve successful grasps, as the proposed control mode requires adequate upward traction to counteract the gravity of the soft hand exoskeleton. This necessary traction is directly proportional to the surface friction coefficient, assuming a constant torque output from the actuators. Objects with lower friction coefficients, such as oranges, chip cans, and coffee cans, present challenges for maintaining sufficient traction, often resulting in unsuccessful grasps. This finding suggests that enhancing the exoskeleton with rough leather patches could improve the GAS by increasing the frictional interface, ensuring a more reliable and effective grip across a wider range of objects.

\begin{figure}[t]
\centering
\includegraphics[width=0.5\textwidth]{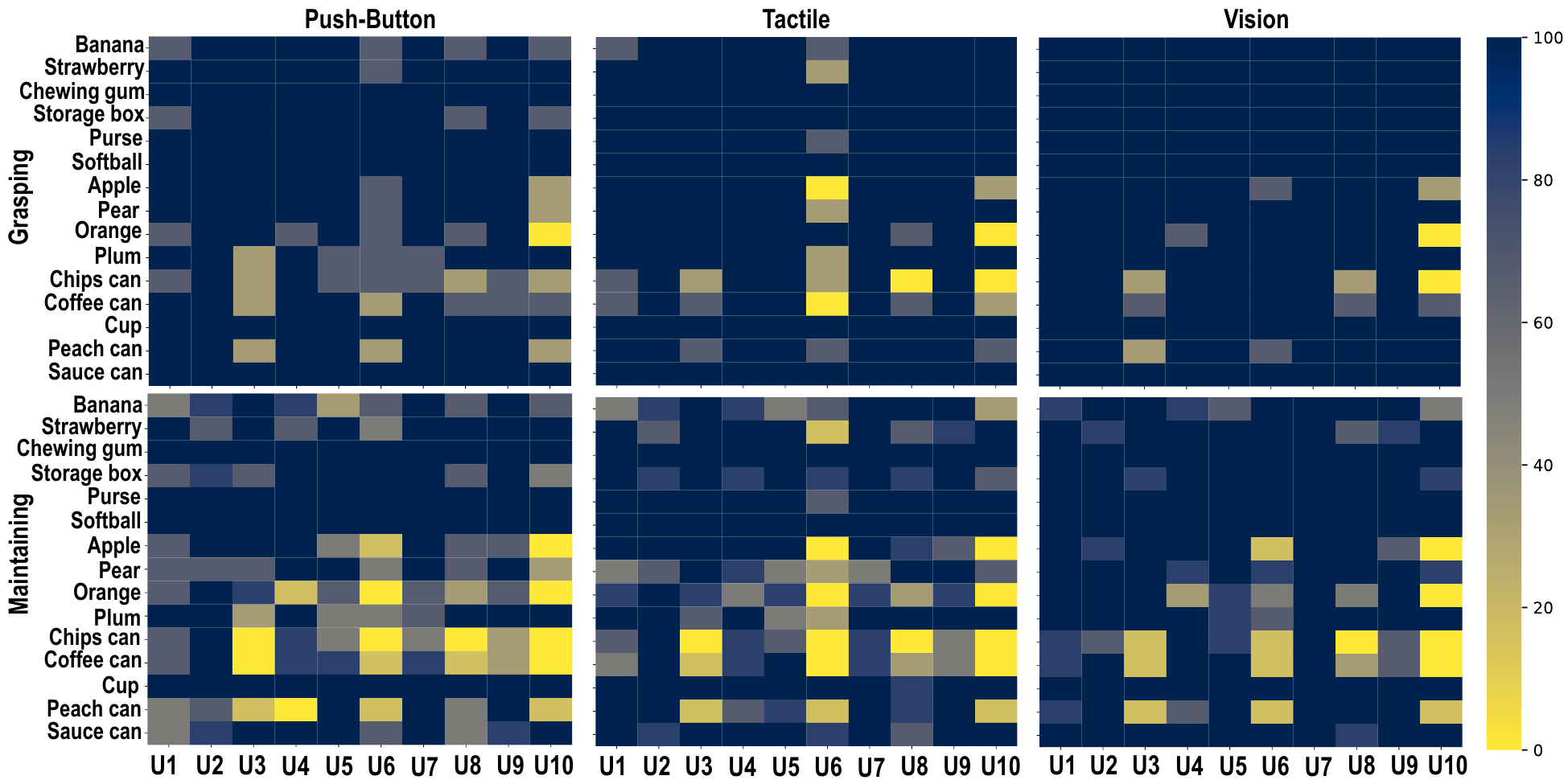}
\caption{Confusion matrix depicting the average Grasping Ability Scores (GAS) across 10 users for 15 objects evaluated in three modes: push-button, force-sensing, and vision. The intensity of the colour reflects the GAS, with darker blue indicating higher scores and lighter yellow indicating lower scores.}
\label{heatmap_grasp_maintain}
\vspace{-0.6cm}
\end{figure}

\begin{figure*}[t]
\centering
\includegraphics[width=0.9\textwidth]{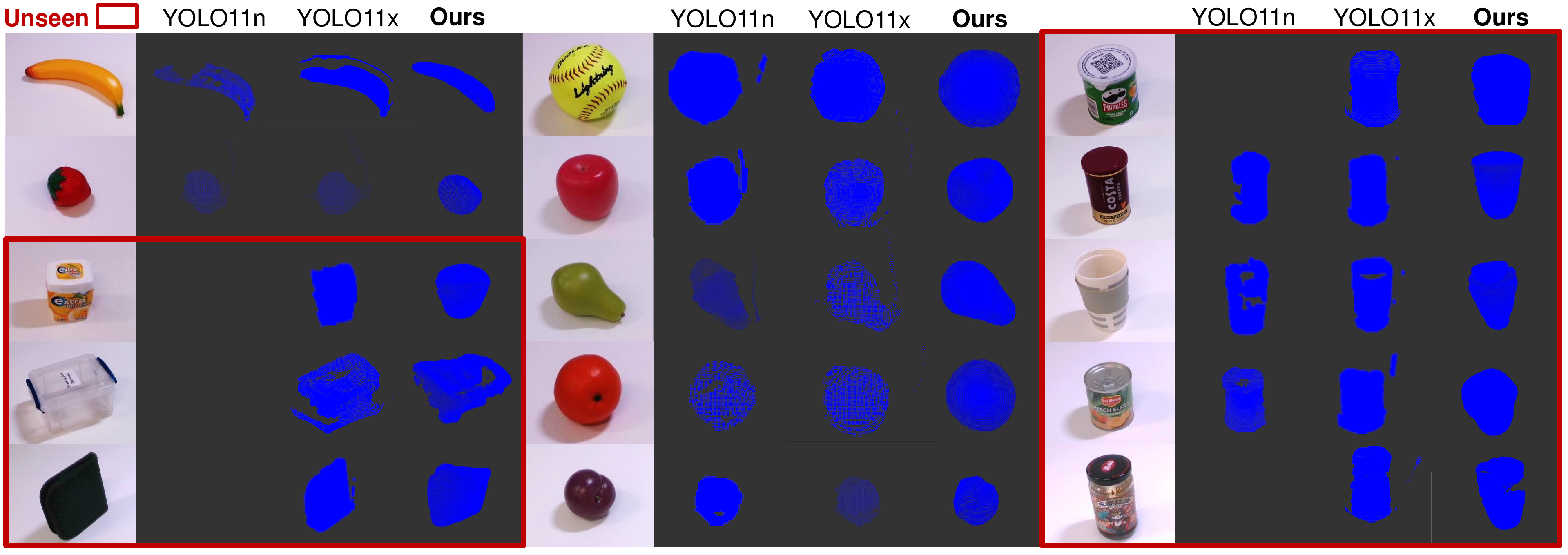}
\vspace{-0.3cm}
\caption{Visual comparison of the proposed method and data-driven approaches in reconstructing 15 objects. Data-driven models (YOLO11n-seg, YOLO11x-seg) struggle with \textit{'unseen'} objects (red boxes), particularly irregular shapes (strawberry, pear), due to segmentation errors under varying viewpoints. Reconstructions also degrade as the camera moves closer. In contrast, our method achieves consistent, complete reconstructions across textit{seen'} and textit{'unseen'} objects, showing generalizability to viewpoint variations without relying on dataset-specific training.}
\label{plot_point_cloud_detection}
\vspace{-0.55cm}
\end{figure*}

The confusion matrix of the GAS for 15 objects assessed by 10 participants across grasping and maintaining tasks shows a distinct pattern where success rates for initial grasping outpace those for maintaining (Fig. \ref{heatmap_grasp_maintain}). Darker colour blocks represent higher GAS. This discrepancy becomes especially evident when participants handle objects with smoother surfaces. Examples include User6 and User10, who can grasp an orange and a coffee but struggle to maintain a stable hold for three seconds after lifting the items. This pattern across all operational modes indicates that while the exoskeleton’s actuation system can provide sufficient torque for effective grasping, it falls short of offering the necessary friction to support prolonged stability.

\begin{table}[t]
\centering
\renewcommand{\arraystretch}{1.5}
\caption{Grasping performance averages divided into Grasping, Maintaining, and total GAS scores for the three types of grasp.}
\label{tab:grasping_performance}
\resizebox{\columnwidth}{!}{%
\begin{tabular}{l l c c c}
\hline
\textbf{Method}        & \textbf{GAS (\%)}      & \textbf{Pinch}            & \textbf{Spherical Grip}   & \textbf{Cylindrical Grip} \\ \hline
\textbf{Push-button}   & Grasping score          & 94.67 $\pm$ 0.72          & 88.00 $\pm$ 1.89         & 84.00 $\pm$ 1.84         \\
~                      & Maintaining score       & 89.33 $\pm$ 1.19          & 74.33 $\pm$ 3.12         & 92.33 $\pm$ 1.25         \\
~                      & GAS score               & 92.00 $\pm$ 0.95          & 81.16 $\pm$ 2.1          & 74.16 $\pm$ 2.29         \\ \hline
\textbf{Force-sensing} & Grasping score          & 96.67 $\pm$ 0.75          & 91.33 $\pm$ 1.89         & 85.33 $\pm$ 1.97         \\
~                      & Maintaining score       & 89.33 $\pm$ 1.59          & 78.00 $\pm$ 2.44         & 91.73 $\pm$ 2.05         \\
~                      & GAS score               & 93.00 $\pm$ 1.17          & 84.66 $\pm$ 2.17         & 78.33 $\pm$ 2.23         \\ \hline
\textbf{Maldonado-Mej\'ia et al.} & Grasping score  & 59.44 $\pm$ 0.26          & 75.33 $\pm$ 0.14         & 93.33 $\pm$ 0.78         \\
~                      & Maintaining score       & 93.33 $\pm$ 0.47          & 57.44 $\pm$ 0.74         & 92.22 $\pm$ 0.34         \\
~                      & GAS score               & 76.39 $\pm$ 0.11          & 83.89 $\pm$ 0.26         & 80.28 $\pm$ 0.31         \\ \hline
\textbf{Proposed approach} & Grasping score      & \textbf{100.00 $\pm$ 0.00} $\uparrow$ & \textbf{95.33 $\pm$ 0.27} $\uparrow$ & \textbf{91.33 $\pm$ 0.55} $\uparrow$ \\
~                      & Maintaining score       & \textbf{95.67 $\pm$ 0.85} $\uparrow$ & \textbf{87.67 $\pm$ 1.72} $\uparrow$ & 76.67 $\pm$ 2.25 \\
~                      & GAS score               & \textbf{97.84 $\pm$ 0.43} $\uparrow$ & \textbf{91.50 $\pm$ 0.59} $\uparrow$ & \textbf{84.00 $\pm$ 0.99} $\uparrow$ \\ \hline
\end{tabular}%
}
\end{table}


Tab. \ref{tab:grasping_performance} compares grasping scores, maintaining scores, and overall GAS across three grip types, pinch, spherical, and cylindrical, for the proposed vision approach\cite{maldonado2023fabric}. 
These results highlight the efficacy of the vision-based method in enhancing grasping performance, particularly in scenarios that require precision (Grasping score) and stability (Maintaining score).

\subsection{Vision-based mode vs. data-driven approaches}
To evaluate the advantages of the proposed method over data-driven approaches, we assess its performance across: computational efficiency and generalizability.

\subsubsection{Computational efficiency and real-time constraints}

Table \ref{tab:system_performance} presents a comparison of CPU processing fps time among the proposed method and two data-driven baselines, YOLO11n-seg and YOLO11x-seg. The proposed approach achieves a processing speed of 10.72 ± 0.58 fps, which is approximately 3.7× faster than YOLO11n-seg, and 39.7× faster than YOLO11x-seg. This substantial improvement highlights the efficiency of real-time applications.


\subsubsection{Generalisation and zero-shot performance}

To assess the generalizability, we evaluate the RSR for both \textit{'seen'} and \textit{'unseen'} objects. As shown in Tab. \ref{tab:system_performance}, data-driven models exhibit varying degrees of generalization failure. YOLO11n-seg achieves 77.14 ± 16.08\% RSR for \textit{'seen'}, but performs poorly (35.00 ± 17.07\%) on \textit{'unseen'} objects, highlighting its limited adaptability. YOLO11x-seg demonstrates slightly improved generalization (77.50 ± 14.94\% on \textit{'unseen'}).

Zero-shot generalization in data-driven models is strongly correlated with model capacity. YOLO11x-seg, with 62.1M parameters, exhibits better generalization than YOLO11n-seg with 2.9M parameters, suggesting that larger models capture more robust feature representations. However, even with an increased parameter count, YOLO11x-seg still fails to generalize reliably beyond the training distribution. In contrast, the proposed method achieves 94.29 ± 3.36\% and 92.50 ± 3.33\% RSR for \textit{'seen'} and \textit{'unseen'} objects, respectively, showing superior performances to novel instances without requiring prior object-specific training.

Another critical limitation of data-driven models is their susceptibility to object geometry and camera viewpoints. Irregularly shaped objects exhibit lower reconstruction success rates depending on their placement angles. Objects such as a strawberry and a pear are correctly reconstructed from a given perspective but fail when the camera angle shifts (Fig. 6), due to segmentation inaccuracies, where the model struggles to correctly delineate object boundaries under previously unseen poses, leading to incomplete or erroneous reconstructions.




\begin{table}[t]
    \centering
    \renewcommand{\arraystretch}{1.5}
    \caption{Comparison of the proposed Visual method and Data-Driven Approaches in CPU Processing Time and Reconstruction Rate.}
    \label{tab:system_performance}
    \resizebox{\columnwidth}{!}{
    \begin{tabular}{lccc}
        \hline
        \textbf{Methods} & \textbf{CPU Processing Time (fps)} & \textbf{Seen (\%)} & \textbf{Unseen (\%)} \\ \hline
        YOLO11n-seg   & 2.89 $\pm$ 0.27   & 77.14 $\pm$ 16.08  & 35.00 $\pm$ 17.07 \\ \hline
        YOLO11x-seg   & 0.27 $\pm$ 0.02   & 80.00 $\pm$ 15.34  & 77.50 $\pm$ 14.94 \\ \hline
        \textbf{Proposed approach}   & \textbf{10.72 $\pm$ 0.58} $\uparrow$  & \textbf{94.29$\pm$ 3.36} $\uparrow$  & \textbf{92.50 $\pm$ 3.33} $\uparrow$ \\ \hline
    \end{tabular}
    }
\vspace{-0.4cm}
\end{table}

\subsection{Kinematics Analysis of Fingers}

\begin{figure}[t]
\centering
\includegraphics[width=0.45\textwidth]{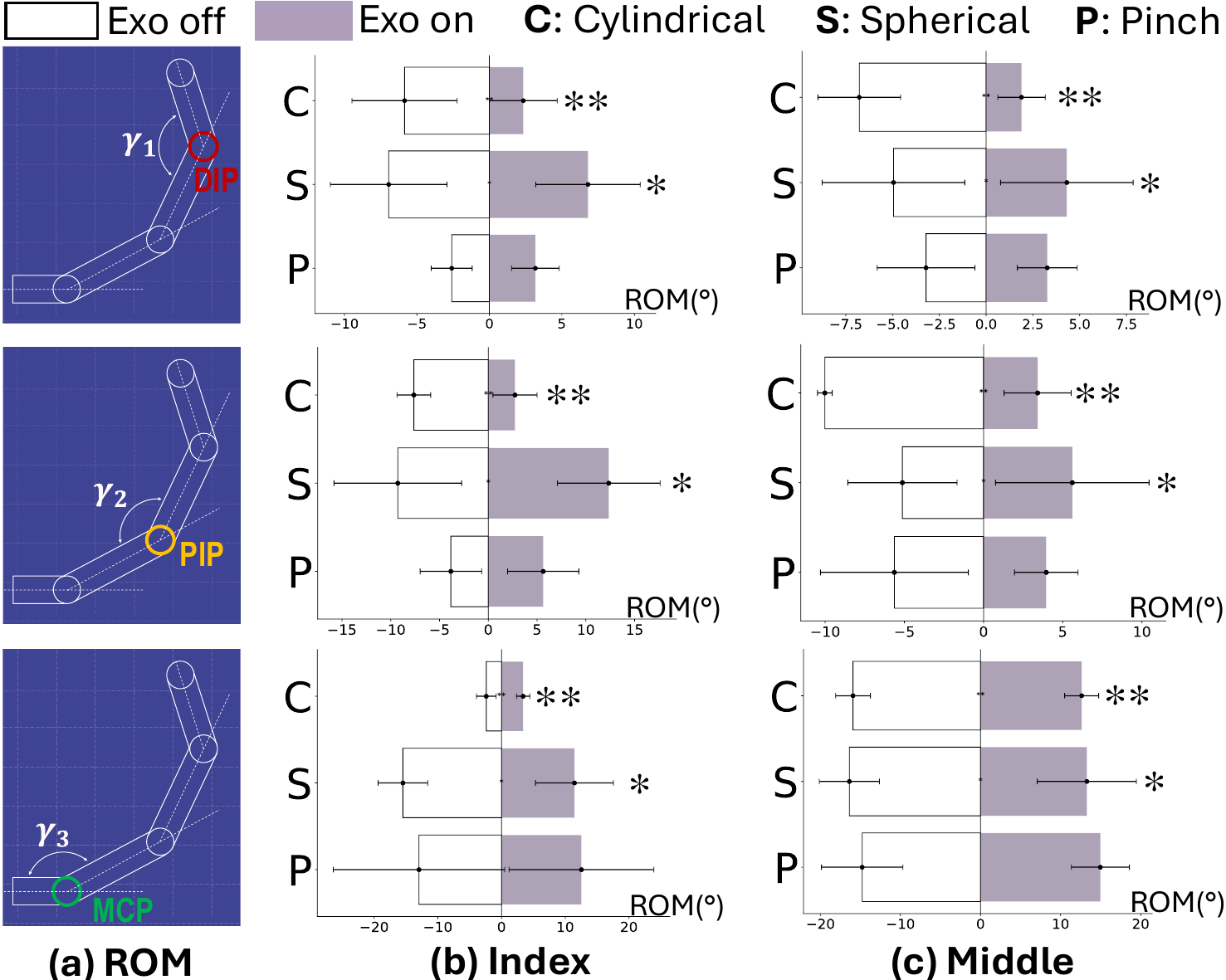}
\caption{Hand kinematics analysis showing the Range of Motion (ROM) with the hand exoskeleton on (shaded bars) and off (unshaded bars) during three grasp types: Cylindrical (C), Spherical (S), and Pinch (P). The ROM is measured for the metacarpophalangeal (MCP), proximal interphalangeal (PIP), and distal interphalangeal (DIP) joints of the index and middle fingers. }
\label{plot_kinematics}
\end{figure}

The impact of the proposed controller on the ROM (Fig. \ref{plot_kinematics}) was consistently higher for the index finger across all grasp types. For cylindrical grasps, the ROM at the MCP joint increased significantly (p $\leq$ 0.01), suggesting improved support from the controller. Similarly, spherical grasps also showed a statistically significant increase in ROM (p $\leq$ 0.05), while pinch grasps exhibited a slight increase. A comparable trend was observed in the middle finger. The cylindrical grasp ROM increased at both MCP and PIP joints (p $\leq$ 0.01), and spherical grasps demonstrated a moderate improvement (p $\leq$ 0.05). The pinch grasp ROM for the middle finger was not significantly affected, maintaining functional consistency. 





\section{Conclusion and Future work}

This study introduced a vision-based control framework tested on a tendon-driven soft hand exoskeleton. The approach can perform contextual perception, target object detection, and relational graph generation for PID control, which achieves a GAS of 91±2\%. Compared to other control strategies, such as push-button, force-sensing, or data-driven methods, the proposed approach increases accuracy and decreases computational complexity. 

Several avenues for future research can further enhance the proposed system. 
Future efforts will focus on developing dynamic actuators able to infer and modulate grasp strength in response to material properties to handle soft or fragile items, improving versatility and safety in various tasks. The system's low computational cost opens the possibility of implementation on embedded devices. The integration of event-based cameras presents an opportunity to address bandwidth-latency trade-offs, inherent to frame-based cameras, by offering high update rates with low bandwidth requirements. 
By exploring these directions, the system has the potential to broaden its functionality, contributing to improving the quality of life for individuals with motor impairments.



\bibliography{references}
\bibliographystyle{IEEEtran}



\end{document}